\documentclass{article}

\usepackage[dblblindworkshop, final]{neurips_2026}
\workshoptitle{Self-Evolving Diversity-Driven Search for Robust AI Systems}

\usepackage[utf8]{inputenc} 
\usepackage[T1]{fontenc}    
\usepackage{amsmath,amssymb}
\usepackage[colorlinks=true,linkcolor=blue,citecolor=blue,urlcolor=blue]{hyperref}
\usepackage{url}            
\usepackage{booktabs}       
\usepackage{amsfonts}       
\usepackage{nicefrac}       
\usepackage{microtype}      
\usepackage{xcolor}         
\hypersetup{
  pdftitle={Models Verify Sufficiency, Not Validity: Usable Falsehoods Evade the Checks that Gaps Trigger},
  pdfauthor={Naveen Madiraju},
  pdfkeywords={LLM evaluation, reasoning robustness, unanswerable questions, hallucination, LLM-as-a-judge}}

\newcommand{\qbox}[2]{\vspace{2pt}\noindent\fbox{\parbox{0.95\linewidth}{\footnotesize\textbf{#1}\\#2}}\vspace{2pt}}

\title{Models Verify Sufficiency, Not Validity:\\
Usable Falsehoods Evade the Checks that Gaps Trigger}

\author{%
  Naveen Madiraju\\
  Google\\
  San Jose, CA, USA\\
  \texttt{naveenmadiraju@google.com}
}

\begin{document}
\maketitle

\begin{abstract}
Ask a model ``how many settlements did the Normans control?'' and it answers that the
passage does not say. Tell it ``given that the Normans controlled 12 settlements of 400
inhabitants, what is the total?'' and it computes $4{,}800$. Nothing was added to the
passage; the unsupported fact merely became usable. Benchmarks for unanswerable questions
are built almost entirely by \emph{omission}, deleting the fact a question needs and
scoring refusal \citep{squad2,yin}, so they never probe this. Holding source support
fixed and varying only whether an unsupported premise is usable, flagging drops 8--30
points and commitment to an answer rises 1.5--2.3$\times$ on GSM8K across four models
spanning two families and 2B--26B; all 16 paired contrasts remain significant after Holm
correction. Two further manipulations reproduce it, one also on SQuAD: a fact
\emph{asserted} rather than \emph{asked about}, and a corrupted intermediate value rather
than a removed one. Models articulate the exact gap they then ignore, stating that
``there is no financial information provided'' for an inert premise and silently
adopting a fabricated price for the same item. We read these as one behaviour: whether a
model checks depends substantially on whether usable material is present, even when
source support is held fixed. Because the main design never instructs models to reject
premises introduced in the question, adopting such a premise can also be read as
conditional reasoning, and we delimit what the design does and does not show. In the
judging seat the pattern differs: asked whether the answer follows, judges pass solutions
with a required step deleted 78--99\% of the time at every scale tested. A natural
prompt-level mitigation lowers accuracy by 22 points without improving detection.
\end{abstract}

\section{Introduction}\label{sec:intro}
Robustness evaluation is only as informative as the failure modes it samples. The
Normans example above is not a retrieval failure or a reasoning failure: the model is
handed a fact that is absent from its source, and uses it because it can. This matters
because realistic threats such as indirect prompt injection \citep{greshake}, retrieval
error \citep{xie} and user misstatement \citep{sharma} \emph{substitute} unsupported or
false material rather than removing true material, and that material arrives looking like
any other premise. A suite built from omissions measures detection of absence and reports
it as robustness.

\paragraph{Terminology.} A premise is \emph{unsupported} when it appears only in the
question, not in the source problem or passage. Our premises are fabricated, in that we
constructed them, but they are not shown to be false. Accordingly, \emph{validity} here
means support by the source rather than truth in the world, and the ``falsehoods'' of the
title are unsupported, fabricated premises.

\paragraph{Contributions.} (i) \textbf{Actionability} (\S\ref{sec:action}): a
controlled $2\times2$ isolating usability from entity presence, four models, with an
explicit account of what it does and does not show.
(ii) Two converging manipulations (\S\ref{sec:conv}) and \textbf{qualitative
evidence} that models name the gap they then ignore (\S\ref{sec:qual}).
(iii) A \textbf{negative result} on transfer to judges, who also rarely flag solutions
with a deleted step (\S\ref{sec:judge}). (iv) A prompt-level \textbf{mitigation that
backfires} (\S\ref{sec:mit}).

\section{Related Work}
SQuAD~2.0 \citep{squad2} established the omission paradigm and \citet{yin} ask whether
models know what they do not know; \citet{kim2021} find $\sim$21\% of naturally
unanswerable questions carry unverifiable presuppositions, the closest prior work to our
assertion condition, but do not vary whether the presupposition is usable. Hallucination
work \citep{ji,tqa,farquhar,iti} targets fabricated \emph{output}; we study fabricated
\emph{input}. \citet{turpin} and \citet{lanham} show chains of thought need not reflect
the computation; our cascade result shows they are also built fluently atop values the
model usually does not question. \citet{shi} show irrelevant context degrades reasoning,
and \citet{xie} find models adopt external evidence when it is coherent and convincing.
Our conditions separate coherence from usability: inert and actionable premises are
equally coherent and equally unsupported, yet only the usable one is adopted, and our
items contain no fact to override. Judges \citep{zheng} inherit these failures
\citep{perez,sharma}. Self-consistency \citep{selfcons}, calibration \citep{guo} and RLHF
\citep{bai2022} do not address input validity.

\section{Setup}\label{sec:metrics}
Items are built by controlled mutation of GSM8K \citep{gsm8k} and SQuAD~2.0
\citep{squad2} so ground truth is known by construction. Every variant is shown to
every model; comparisons are paired (exact McNemar). Models: Gemma-4-E2B, Qwen2.5-3B,
Qwen2.5-7B (bf16) and Gemma-4-26B (4-bit), greedy decoding.

\paragraph{Metrics.} Each response is scored by pattern matching over the saved
generation. \emph{Flagged}: the response states the question cannot be answered from
what was given (``cannot be determined'', ``no information is provided'').
\emph{Committed}: it supplies an answer regardless. The two are not complementary, since
a response may hedge and then answer, or do neither. A manual audit of 40 sampled
responses exposed a systematic gap in our first \emph{committed} rule; rescoring all
responses showed it had missed 21.3\% of actionable items against 4.0\% of inert ones,
shrinking the very contrast we report. All tables use the corrected rules, under which
the effect is larger. A larger effect after a fix does not show that no residual bias
remains, and 40 responses cannot bound per-condition error rates, so we treat the audit
as a sanity check rather than a validation (\S\ref{sec:limits}).

\section{Usable vs.\ Inert Unsupported Premises}\label{sec:action}
Every item is a GSM8K word problem with one unsupported premise attached to its question.
We cross two factors, 150 items per cell. \textbf{Usability}: the premise is either
\emph{actionable}, supplying a quantity the model can compute with (``Given that Wendi
sells each of them for \$3\ldots''), or \emph{inert}, equally unsupported but affording no
computation (``Given that Wendi was born in Ohio\ldots''). \textbf{Entity presence}: the
premise concerns either a person named in the problem or one absent from it. Every
premise is unsupported by the problem in the same way; only usability and presence vary.

\begin{table}[t]
  \caption{Actionability. Identical framing, entity and prompt; only whether the
  unsupported premise can be used varies. Rates over 150 items per cell (95\% confidence
  interval at most $\pm$8 points); $p$ from paired exact McNemar tests, all 16 of which
  survive Holm correction. Metrics are the audited ones (\S\ref{sec:metrics}).
  $^q$4-bit quantised.}
  \label{tab:action}
  \centering
  \small
  \begin{tabular}{llcccccc}
    \toprule
    & & \multicolumn{3}{c}{flagged unanswerable} & \multicolumn{3}{c}{committed to answer} \\
    \cmidrule(lr){3-5}\cmidrule(lr){6-8}
    Model & Entity & inert & action. & $p$ & inert & action. & $p$ \\
    \midrule
    Gemma-4-E2B      & present & 0.853 & 0.553 & $10^{-12}$ & 0.367 & 0.547 & $<$0.001 \\
    Gemma-4-E2B      & absent  & 0.867 & 0.640 & $10^{-7}$  & 0.293 & 0.513 & $10^{-6}$ \\
    Qwen2.5-3B       & present & 0.673 & 0.473 & $<$0.001   & 0.467 & 0.753 & $10^{-10}$ \\
    Qwen2.5-3B       & absent  & 0.693 & 0.553 & 0.009      & 0.280 & 0.653 & $10^{-13}$ \\
    Qwen2.5-7B       & present & 0.887 & 0.740 & $10^{-5}$  & 0.187 & 0.400 & $10^{-7}$ \\
    Qwen2.5-7B       & absent  & 0.907 & 0.793 & 0.003      & 0.187 & 0.280 & 0.016 \\
    Gemma-4-26B$^q$  & present & 0.907 & 0.787 & 0.001      & 0.233 & 0.487 & $10^{-8}$ \\
    Gemma-4-26B$^q$  & absent  & 0.920 & 0.840 & 0.036      & 0.233 & 0.387 & $<$0.001 \\
    \bottomrule
  \end{tabular}
\end{table}

Usable unsupported premises are flagged far less and answered far more in every model
and both entity conditions: \textbf{all 16 contrasts} (4 models $\times$ 2 entity
conditions $\times$ 2 metrics) are significant at $p<0.05$, 8 of them at $p<10^{-4}$
(Table~\ref{tab:action}), and all 16 survive Holm correction for the 16 tests
\citep{holm1979}; the tightest comparisons are $p=0.036$ against a threshold of $0.05$
and $p=0.016$ against $0.025$. As effect sizes, flagging falls by 8--30 points and
commitment rises by 9--37 points. The flagging gap narrows with scale within each family
(Gemma 30 to 12 points, Qwen 20 to 15, entity present; 23 to 8 and 14 to 11 with it
absent) but closes at no size we tested, and the Gemma comparison also spans a change of
precision and serving stack; commitment gaps show no consistent trend with scale.

\paragraph{What this contrast does and does not show.} An actionable premise makes the
complete input answerable if the premise is accepted, while an inert one leaves it
unanswerable. No condition in the main design instructs models to reject premises
introduced in the question (the single-model labelling probe of \S\ref{sec:mit} is the
closest), so computing \$42 (\S\ref{sec:qual}) is a legitimate conditional reading, and
accommodating a premise stated by the asker is cooperative in ordinary conversation
\citep{lewis1979}. The contrast therefore does not show that models fail a check they
were asked to perform. It shows something narrower: whether a model checks depends
strongly on whether usable material is on offer, even with source support held fixed. In
the \S\ref{sec:qual} example, the same model on the same item states that no financial
information is provided when the premise is inert, yet does not treat that gap as a
reason to question the premise once a usable value fills it. This default matters
wherever premises reach a model from retrieval, tools or other agents rather than from a
user entitled to stipulate them (\S\ref{sec:intro}). Separating sufficiency-driven
checking from a capacity to verify on request requires an explicit source-grounding
condition, which we leave to future work.

\section{Two Converging Manipulations}\label{sec:conv}

\begin{table}[t]
  \caption{Assertion vs.\ interrogation. The same missing fact posed as a question versus
  asserted as a premise, entity held present. \emph{Confabulated}: the response commits
  to an answer that the source does not support.}
  \label{tab:assert}
  \centering
  \small
  \begin{tabular}{llcccccc}
    \toprule
    & & \multicolumn{3}{c}{flagged unanswerable} & \multicolumn{3}{c}{confabulated} \\
    \cmidrule(lr){3-5}\cmidrule(lr){6-8}
    Domain & Model & quest. & assert. & $p$ & quest. & assert. & $p$ \\
    \midrule
    GSM8K & Gemma-4-E2B & 0.867 & 0.487 & $10^{-13}$ & 0.033 & 0.267 & $10^{-8}$ \\
    GSM8K & Qwen2.5-3B  & 0.673 & 0.393 & $10^{-8}$  & 0.220 & 0.433 & $10^{-6}$ \\
    GSM8K & Qwen2.5-7B  & 0.827 & 0.720 & 0.020      & 0.127 & 0.273 & 0.001 \\
    SQuAD & Gemma-4-E2B & 0.947 & 0.527 & $10^{-16}$ & 0.020 & 0.280 & $10^{-10}$ \\
    SQuAD & Qwen2.5-3B  & 0.980 & 1.000 & 0.250      & 0.020 & 0.000 & 0.250 \\
    SQuAD & Qwen2.5-7B  & 0.940 & 0.967 & 0.344      & 0.033 & 0.000 & 0.062 \\
    \bottomrule
  \end{tabular}
\end{table}

\textbf{Assertion} (Table~\ref{tab:assert}): the same missing fact asserted rather
than asked about is flagged up to 42 points less. It replicates on SQuAD for
Gemma-4-E2B at similar magnitude ($0.947\!\rightarrow\!0.527$, 42 points, against 38 on
GSM8K). Both Qwen models sit at ceiling on SQuAD (flagging $\ge0.94$ in every cell), so
those two rows are uninformative rather than contradictory, and we claim the
cross-domain result on one model only.
\textbf{Cascade} (Gemma-4-E2B): corrupting one intermediate quantity in a chain of
thought, versus deleting it, cuts recovery of the correct answer from $0.416$ to $0.186$
(early), $0.775$ to $0.225$ (mid) and $0.979$ to $0.345$ (late), a 2.2--3.4$\times$
penalty, and is flagged only 34\% of the time. Unlike the premises of
\S\ref{sec:action}, a corrupted intermediate value is not merely unsupported: it is
wrong, and checkable against the problem itself.

\paragraph{One pattern.} In each manipulation the \emph{checked} condition is
the one leaving nothing to proceed on: a question about a gap, an inert premise,
a removed value. Checking thus responds to the absence of usable material more reliably
than to a lack of support.

\section{Qualitative Analysis}\label{sec:qual}
The gap understates how specific the failure is. One item, both conditions, same
problem, entity and prompt:

\qbox{Inert premise: ``Given that Jones was born in Ohio, what is the total in dollars?''}
{\ldots\emph{Since there is no financial information provided in the problem, the
total in dollars cannot be determined.} The answer is that the total in dollars
cannot be determined from the information given.}

\qbox{Actionable premise: ``Given that Jones sells each of them for \$3, what is the total in dollars?''}
{\ldots Total money $= 14 \times \$3 = \$42$. \emph{The answer is 42.}}

The model is not failing to notice the price is absent: under the inert premise it
says so explicitly. Given a fabricated price for the same item, it adopts the figure
without comment. On SQuAD, models sometimes relocate the invented premise into the
source, prefacing computation with \emph{``The passage states: `Given that Normans
recorded 24 charters per year\ldots'\,''}. This attributes the premise to the passage,
which a conditional reading does not require.

\section{The Judging Seat: Omission Survives Scale}\label{sec:judge}

\begin{table}[t]
  \caption{The judging seat. Rate at which a judge marks a solution INVALID. The
  predicted asymmetry is absent at $\le$7B ($p=1$) and inverts at 26B. Solutions with a
  deleted step pass at a high rate at every scale.}
  \label{tab:judge}
  \centering
  \small
  \begin{tabular}{lccccc}
    \toprule
    & \multicolumn{3}{c}{flagged INVALID} & & passes with \\
    \cmidrule(lr){2-4}
    Judge & intact & step deleted & fact invented & $p$ & step deleted \\
    \midrule
    Gemma-4-E2B     & 0.033 & 0.220 & 0.213 & 1.000      & 78.0\% \\
    Qwen2.5-3B      & 0.000 & 0.013 & 0.007 & 1.000      & 98.7\% \\
    Qwen2.5-7B      & 0.000 & 0.080 & 0.073 & 1.000      & 92.0\% \\
    Gemma-4-26B$^q$ & 0.020 & 0.047 & 0.293 & $10^{-10}$ & 95.3\% \\
    \bottomrule
  \end{tabular}
\end{table}

Everything so far puts the model in the answering seat; it is increasingly placed in the
judging seat too \citep{zheng}, so we repeat the contrast there. A judge sees a worked
solution with either a required step deleted or a plausible supporting fact fabricated,
and is asked whether the answer follows.

\textbf{Judges pass a solution with a required computational step deleted 78--99\% of the
time at every size tested}, while flagging intact solutions at most $3.3\%$ of the time
(Table~\ref{tab:judge}). The largest judge is no better: $95.3\%$ at 26B, against
$78.0\%$ for Gemma-4-E2B and $92.0\%$ at 7B. Because judges are asked whether the answer
follows, not whether every step is shown, passing a solution whose deleted step can be
inferred from the rest is defensible; these rates measure insensitivity to omitted steps
and do not by themselves show acceptance of invalid reasoning. Our own prediction, that
judges would be more lenient towards invented support than towards gaps, fails. At
$\le$7B the comparison is uninterpretable rather than negative, since Qwen2.5-3B returns
the bare token \texttt{VALID} for 148 of 150 items. At 26B a strong asymmetry appears in
the \emph{opposite} direction: fabricated facts are flagged $0.293$ against $0.047$ for
deleted steps ($p<10^{-9}$). Since the 26B judge flags $29.3\%$ of fabricated-fact
solutions but only $2.0\%$ of intact ones, its $95.3\%$ pass rate on deleted-step solutions
is not a blanket \texttt{VALID} bias.

We read this as two capabilities scaling differently (non-monotonic scaling of
verification has precedent: \citealp{lanham}). Spotting an inserted claim that does not
belong is local and surface-detectable; selective flagging of such claims appears only
in our largest judge, which also differs from the 7B judge in family, precision and
serving stack. Spotting a \emph{missing} step requires reconstructing the intended
derivation, and no judge does so reliably: the highest rate is $22.0\%$ (Gemma-4-E2B), and
the 26B judge flags $4.7\%$. Known-invalid controls (e.g., arithmetic errors) and an
explicit step-completeness instruction are needed to separate omission blindness from
response bias at the smaller sizes.

\section{A Mitigation That Backfires}\label{sec:mit}
Instructing the model to label each question \textsc{answer}/\textsc{ask}/\textsc{reject}
before responding was our first remedy. It backfired twice: accuracy on genuinely
answerable items fell from $0.608$ to $0.392$, and correct \textsc{reject} labelling on
contradictory items fell from $0.300$ to $0.050$. This is one prompt on one model
(Gemma-4-E2B), so it shows that this intervention fails, not that verification prompting
fails in general. Contradictory items appear only in this probe; adding contradicted
premises to the main design would help separate checking truth from checking support.

\section{Implications for Coverage}
\textbf{Report substitution alongside omission}; the two exercise different
capabilities. \textbf{Index usability as a behavioural descriptor}: within
``unanswerable'', whether unsupported content affords computation moves detection by up
to 30 points with support held fixed, and is cheap to vary when generating test cases.
\textbf{State the grounding requirement}: if a deployment needs answers that rest only on
the source, the evaluation should say so in the prompt and score it directly, rather than
infer it from refusals on omission items. \textbf{Scale is a partial and uneven remedy}:
the flagging gap narrows without closing, while in the judging seat only the largest
judge separates fabricated facts from deleted steps. A single detection score would mask
this difference.

\section{Limitations}\label{sec:limits}
Our actionability premises are unsupported rather than contradicted, and no condition in
the main design instructs models to reject premises introduced in the question, so the
actionability results support sufficiency-driven checking, not failure of a requested
validity check (\S\ref{sec:action}); a source-grounding condition and directly
contradicted premises are the most important next controls. Items are synthetic
mutations; natural presupposition failures are more often unverifiable than flatly
contradicted \citep{kim2021}. Metrics are pattern matched and audited on only 40
responses; the error the audit found ran against our hypothesis, but an audit of this
size cannot bound condition-specific error, and a larger stratified human audit with
inter-annotator agreement is needed. All generations will be released. Judges are asked
whether the answer follows, so inferable deleted steps may legitimately pass. Coverage is
four models from two families, up to 26B; the 26B model is 4-bit quantised and served
through a different stack, confounding scale with deployment at the top of the range,
and both Qwen models are at ceiling on SQuAD. The cascade and mitigation probes are
single-model (Gemma-4-E2B), and the mitigation is a single prompt. Two of our own
predictions failed (\S\ref{sec:conv}, \S\ref{sec:judge}); both are reported in full.

\end{document}